\documentclass[12pt]{iopart}
\usepackage{graphicx}
\usepackage{subfig}
\usepackage{float}
\usepackage{cases}
\usepackage{algorithm} %format of the algorithm
\usepackage{algorithmic} %format of the algorithm

\usepackage{latexsym}

\usepackage{url}
\usepackage{xcolor}

\usepackage{hyperref}

\usepackage{amssymb}

\usepackage{bm}
\usepackage{booktabs}
\usepackage{array}

\usepackage{multirow} %multirow for format of table

\begin{document}

\title[AUNet for breast mass segmentation in whole mammograms]{AUNet: Attention-guided dense-upsampling networks for breast mass segmentation in whole mammograms}

\author{Hui Sun${^{1, 2,\dag}}$,Cheng Li${^{1,\dag}}$, Boqiang Liu${^2}$, Zaiyi Liu${^3}$, Jianhua Luo${^4}$, Hairong Zheng${^1}$, David Dagan Feng${^5}$, and Shanshan Wang$^{1,*}$}

\address{$^1$Paul C. Lauterbur Research Center for Biomedical Imaging, Shenzhen Institutes of Advanced Technology, Chinese Academy of Sciences, Shenzhen, Guangdong 518055, China. \\
$^2$School of Control Science and Engineering, Shandong University, Shandong 250100, China.\\
$^3$Department of Radiology, Guangdong General Hospital, Guangdong Academy of Medical Sciences, Guangzhou, Guangdong 510080, China.\\
$^4$School of Aeronautics and Astronautics, Shanghai Jiao Tong University, Shanghai 200240, China.\\
$^5$The Biomedical and Multimedia Information Technology Research Group, School of Information Technologies, The University of Sydney, Sydney, NSW 2006, Australia.\\
~\\
$\dag$ These authors contribute equally to this paper.\\
$*$ Correspondence to Shanshan Wang.}

\ead{sophiasswang@hotmail.com}

\vspace{10pt}
\begin{indented}
\item[]July 2019
\end{indented}

\begin{abstract}
Mammography is one of the most commonly applied tools for early breast cancer screening. Automatic segmentation of breast masses in mammograms is essential but challenging due to the low signal-to-noise ratio and the wide variety of mass shapes and sizes. Existing methods deal with these challenges mainly by extracting mass-centered image patches manually or automatically. However, manual patch extraction is time-consuming and automatic patch extraction brings errors that could not be compensated in the following segmentation step. In this study, we propose a novel attention-guided dense-upsampling network (AUNet) for accurate breast mass segmentation in whole mammograms directly. In AUNet, we employ an asymmetrical encoder-decoder structure and propose an effective upsampling block, attention-guided dense-upsampling block (AU block). Especially, the AU block is designed to have three merits. Firstly, it compensates the information loss of bilinear upsampling by dense upsampling. Secondly, it designs a more effective method to fuse high- and low-level features. Thirdly, it includes a channel-attention function to highlight rich-information channels. We evaluated the proposed method on two publicly available datasets, CBIS-DDSM and INbreast. Compared to three state-of-the-art fully convolutional networks, AUNet achieved the best performances with an average Dice similarity coefficient of 81.8\% for CBIS-DDSM and 79.1\% for INbreast.
\end{abstract}

%
% Uncomment for keywords
%\vspace{2pc}
%\noindent{\it Keywords}: XXXXXX, YYYYYYYY, ZZZZZZZZZ
%
% Uncomment for Submitted to journal title message
%\submitto{\JPA}
%
% Uncomment if a separate title page is required
%\maketitle
% 
% For two-column output uncomment the next line and choose [10pt] rather than [12pt] in the \documentclass declaration
%\ioptwocol
%

\section{Introduction}
Latest investigations demonstrate that breast cancer persists as one of the most threatening cancer types to female, accounting for 29\% of cancer incidence and 15\% of cancer mortality in women \cite{Siegel2017}. Early diagnosis of breast cancer is vital for the survival of patients. Mammography is one of the most effective and efficient breast cancer screening tools. However, analyzing mammograms by radiologists is tedious and the interpretations are subject to substantial inter- and intra-observer variations, which may lead to missed cancers as well as overdiagnosis \cite{Birdwell2001,Loberg2015}. Therefore, a computer-aided detection/diagnosis (CAD) system that can work as a second reader is important and necessary.

Various types of abnormalities may show in mammograms, such as asymmetrical breast tissues, adenopathy, density, microcalcifications, and masses. Among them, breast masses are believed to contribute significantly to breast cancers \cite{Giger2013}. Currently, the majority of breast mass studies concentrated on image-level lesion detection and patch-level mass classification or segmentation \cite{Dhungel2015b,Dhungel2017,Han2017,Jiang2016,Kim2018,Wei2006}. However, image-level lesion detection can only give the bounding box of the mass without the boundary information, which has been identified as an important indicator of its malignancy \cite{Guliato2008}. And patch extraction around the mass before segmentation is a tedious and difficult work for radiologists. Therefore, mass segmentation of whole mammograms is of high application value for breast cancer detection and diagnosis. Specifically, our focus in this study is the automatic breast mass segmentation in whole mammograms, i.e., the segmentation in full fields of view (FOVs) of input mammograms rather than extracted regions of interest (ROIs).

Recently, deep learning models, especially convolutional neural networks (CNNs), have seen great successes in computer vision and medical imaging \cite{Greenspan2016,Hamidinekoo2018,Litjens2017}. In respect of medical image segmentation, the most well-known network is UNet \cite{Ronneberger2015} and UNet-like architectures are frequently investigated \cite{Balagopal2018,Li2019}. However, most deep learning-based models developed for mammographic mass segmentation focus on extracted patches instead of the original whole mammograms \cite{Jinjin2019}. In addition to the limited studies, existing deep learning-based studies conduct whole mammographic mass segmentation by simply combing classic models with some effective network modules developed for natural image processing. Atrous spatial pyramid pooling and attention gates have been introduced to FCDenseNet and Dense-U-Net to enhance the segmentation capacity \cite{Jinjin2019,Li2019b}. In these studies, both the network architecture and the added modules were not specifically optimized for the breast mass segmentation purpose. There is still a large gap to be filled and a lot of work to be done. Besides, although CAD systems have been widely developed to assist radiologists in identifying suspicious regions, their performance can still be improved since contradictory conclusions exist regarding their effectiveness in mammogram interpretation \cite{Lehman2015,Kooi2017}. Therefore, we feel motivated to investigate the whole mammographic mass segmentation project, which is expected to be a significant add-on to the current CAD system for mammographic diagnosis.

In this paper, we propose a new model, attention-guided dense-upsampling network (AUNet), for the segmentation of mammographic masses. Different from the classical symmetric encoder-decoder architecture of UNet \cite{Ronneberger2015}, AUNet employs an asymmetrical structure – different encoder and decoder blocks – through the implementation of residual connections. Furthermore, we design a novel upsampling module, attention-guided dense-upsampling block (AU block), to compensate the information loss caused by bilinear upsampling, effectively fuse the high- and low-level features, and at the same time, highlight the rich-information channels. The performance of the proposed network was evaluated on two public mammographic datasets, CBIS-DDSM and INbreast. With AUNet, we achieved an average Dice score of 81.8\% for CBIS-DDSM and 79.1\% for INbreast. Both improved the segmentation results of UNet by more than 8\%. Our major contributions are: 1) A more effective asymmetric encoder-decoder network architecture is introduced; 2) We propose a new block, AU block, that can effectively extract important information from both high- and low-level features; 3) AU block can serve as a universal decoder module that is compatible with any encoder-decoder segmentation network; 4) Implementing both AU block and the asymmetrical structure, our proposed network, AUNet, is able to accurately segment masses in whole mammograms without the need of ROI extraction; 5) Superior breast mass segmentation performances were achieved by AUNet compared to commonly utilized fully convolutional networks (FCNs) in medical imaging. Our code will be made publicly available soon.

\section{Related works}
In this section, we review the related works on deep learning models for image segmentation and existing methods for mammographic mass segmentation.

\subsection{Segmentation networks}
Since the introduction of FCNs in 2015 \cite{Long2015}, most segmentation models follow a similar encoder-decoder network backbone design. The encoder pathway first extracts high dimensional and high abstract feature maps from the inputs, usually with severely decreased resolutions, and then the decoder pathway is responsible for the recovery of image resolution and generation of the segmentation results. However, due to the information loss during the encoding process by pooling or convolution with strides, the reconstructed segmentation results are usually not satisfactory. To solve this issue, works have been done to include conditional random fields as a post processing method, which has shown a significant improvement \cite{Chen2018,Kamnitsas2017}. Another direction is the application of dilated convolution \cite{Yu2017}. Dilated convolution can increase the receptive field and, in the meantime, keep the image resolution unchanged. Nevertheless, limited by the current available computing power, dilated convolution at high image resolutions is hard to achieve if not impossible \cite{Yu2017}. UNet proposed another solution to the problem \cite{Ronneberger2015}. The main idea of UNet is to fuse high-level feature maps that are rich in semantic information with low-level feature maps that are rich in location information. By fusing feature maps from different layers, UNet is capable of generating accurate segmentation maps for small datasets. However, the feature fusion of UNet is done through simple concatenation, which is not effective enough and improvement is necessary for different applications \cite{Lin2017,Zhang2018}.

\subsection{Upsampling approaches}
Different methods have been adopted in literature to upsample the low-resolution feature maps. Bilinear interpolation is a simple and efficient method that has been commonly used \cite{Chen2018,Zhao2017}. The output of bilinear interpolation is fixed and not learnable, which may cause information loss \cite{Wang2018}. Deconvolution was first proposed along with FCNs \cite{Long2015} and adopted in later works. Deconvolution could be realized in two ways. One is through the reverse operation of convolution \cite{Long2015}. The other is through unpooling, where the low-resolution feature maps are first upsampled to high-resolution feature maps using the stored max pooling indices and then the sparse feature maps are densified by convolutions \cite{Badrinarayanan2017}. Both methods result in learnable upsampling procedure but require zero padding at the first step. The last method is dense upsampling convolution (DUC) \cite{Wang2018}, derived from the sub-pixel convolution method originally developed for image super resolution task \cite{Shi2016}. DUC is also learnable. In addition, different from deconvolution, no zero padding is required for DUC.

\subsection{Attention mechanism}
Attention mechanism in neural networks has attracted a lot of attention recently. It is proposed in accordance with the human visual attention that human beings always focus on a certain part of a given image after quickly glimpsing through it. Attention could be viewed as a tool to force the network focusing on the most informative part of the inputs or features \cite{Mnih2014}. It has been widely applied in natural language processing and image captioning \cite{Chen2017,Vaswani2017}. Studies also found that CNNs could learn implicitly to localize the most important regions of the input images \cite{Zhou2016}, which could be treated as a kind of attention. To improve image classification accuracies, both spatial and channel-wise attention modules have been proposed in literatures \cite{Hu2018,Roy2018}. Attention has also been explicitly used for image segmentation \cite{Mirikharaji2018,Nie2018}. Different from these works, which utilize attention mechanism to focus on regions of inputs, our proposed AU block implements attention to select important channels for breast mass segmentation.

\subsection{Segmentation of mammographic mass}
Automatic mammographic mass segmentation methods could be divided into unsupervised and supervised methods. Unsupervised methods include region-based \cite{Gulsrud2005,Wei2006}, contour-based \cite{Rahmati2012,Shi2008}, and clustering models \cite{Abdeldayem2005,Ball2004}. These models encounter various problems when applied to mammographic mass segmentation \cite{Oliver2010}. Region-based models rely on region homogeneity and prior information is usually needed, such as the locations of seeding points and shape information \cite{Kupinski1998}. Contour-based models are based on edge detection whereas it is challenging to extract the boundary between masses and normal breast tissues \cite{Sahiner2001}. Hierarchical clustering models are computational expansive while partitional clustering models need to know the number of regions in advance \cite{Li2002}. Supervised methods have a training and testing procedure. Pattern matching is widely used for segmentation and detection \cite{Freixenet2008,Song2010}. Nonetheless, mammographic masses can be in a wide variety of shapes, which hinders the usage of pattern matching approaches \cite{Oliver2010}. Deep learning models belong to supervised methods. Deep structured models have been successfully applied to segment masses from ROIs rather than whole mammograms \cite{Dhungel2015b,Dhungel2017,Dhungel2015c}. And using manually extracted ROIs could improve the segmentation performance compared to automatically detected bounding boxes generated by detection models \cite{Dhungel2017}, which indicates that the segmentation results depend on the patch extraction process and it is difficult to achieve fully automatic mammographic mass segmentation employing this approach. Very few attempts on mass segmentation of whole mammograms could be found probably caused by the previously discussed difficulties \cite{Jinjin2019}. These studies mainly combined famous segmentation models with some special network modules developed for natural image analysis. For example, atrous spatial pyramid pooling and attention gates have been introduced to FCDenseNet and Dense-U-Net to enhance the segmentation capacity \cite{Jinjin2019,Li2019b}. Considering the gap between medical and natural image domains, these models may not be perfectly suitable for the breast mass segmentation task. Moreover, the experiments were not comprehensive, and the models were not publicly available. Aiming to address these challenges, our AUNet is designed specifically for fully automatic mammographic mass segmentation. Two public datasets have been tested and the models will be made available once the paper is accepted.

\section{Methodology}
In this section, we first describe the datasets used in the study. Then, the proposed network architecture, including the asymmetrical encoder-decoder backbone and the AU block, is presented. After that, loss function selection is discussed. Finally, quantitative evaluation metrics are listed.

\subsection{Datasets}
We instantiated our proposed network with two publicly available datasets, CBIS-DDSM \cite{Heath2000,Lee2017} and INbreast \cite{Moreira2012}. For CBIS-DDSM, a total of 858 images were used in the current study with 690 images for training and 168 for validation. The INbreast dataset contains 107 images with accurate mass segmentation masks. A 5-fold cross-validation experiment was conducted for INbreast.

All the images along with the masks were first processed to remove the irrelevant background regions (rows and columns have negligible maximum intensities) and then resized to $256 \times 256$, followed by an intensity normalization. Before inputting into the networks, the gray images were changed to RGB images by copying the pixel values to the other two channels. The importance of this step will be discussed later. No further data processing or augmentation was applied.

\begin{figure}[!t]
\centering
\includegraphics[scale=0.98]{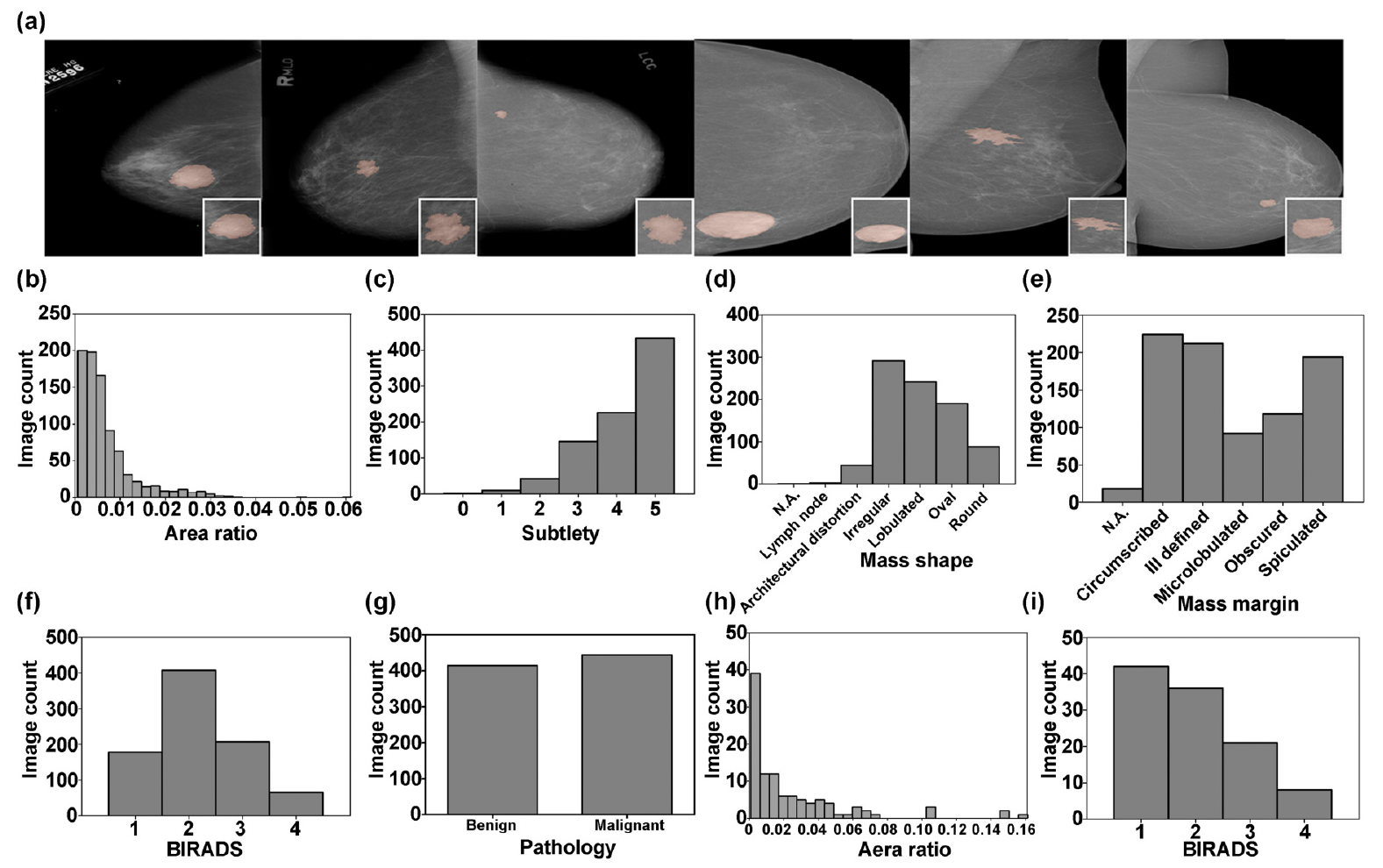}
\caption{Representative mammographic images (a) and distributions of the area ratios (area ratio = area of the mass/area of the whole mammogram) of masses, the subtlety, the mass shape, the mass margin, the BIRADS category, and the pathology diognosis for the CBIS-DDSM dataset (b-g) and area ratios of masses and the BIRADS category for the INbreast dataset (c). The left three images in (a) are from the CBIS-DDSM dataset and the right three from the INbreast. Pink color regions indicate the masses. Insets are enlarged patches contain the masses. (For interpretation of the references to color in this figure legend, the reader is referred to the web version of this article.)}
\label{fig01}
\end{figure}

Fig.\ref {fig01}a shows representative images from the two datasets. It could be observed that mammographic masses are in a wide variety of shapes and sizes, which increases the difficulty of training the segmentation network. Fig. \ref {fig01}b and h give the area ratio distributions of the two datasets. Both indicate that most masses only occupy very small regions of the whole mammograms. Results confirm more than 81.8\% masses occupy less than 1\% area of the whole mammograms for CBIS-DDSM. For INbreast, more than 81\% masses occupy less than 4\% area of the whole mammograms. Therefore, it is much more difficult to train a network capable of accurately segmenting masses in whole mammograms than in mass-centered mammographic patches. Other available important information including subtlety, mass shape and margin, BIRADS category, and pathology are also plotted in Fig.1 to comprehensively describe the datasets.

\subsection{Asymmetrical network backbone}
\begin{figure}[!t]
\centering
\includegraphics[scale=0.5]{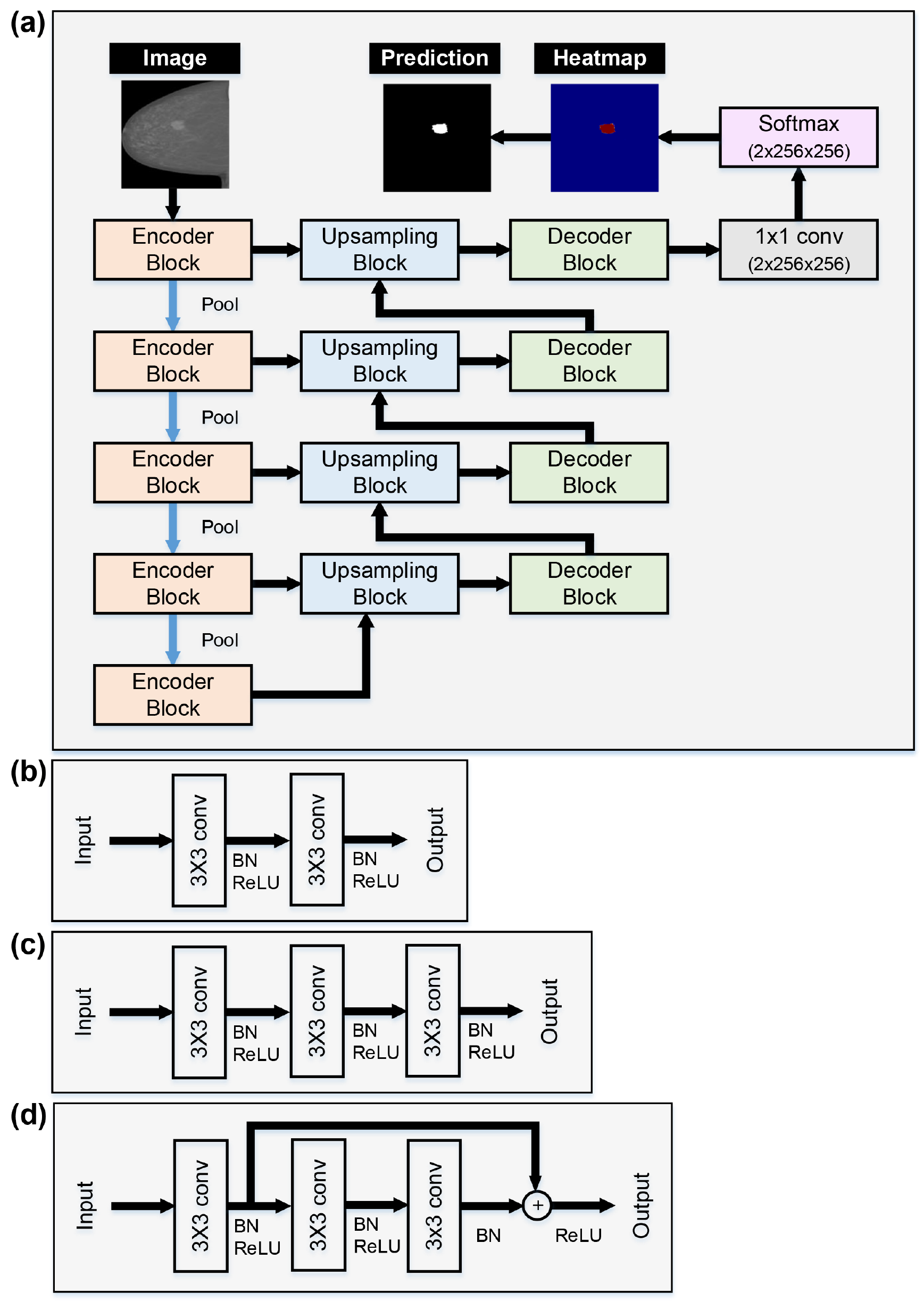}
\caption{Network architecture and different blocks. (a) Architecture backbone. (b)-(d) Different blocks named as basic unit (b), deep unit (c), and res unit (d).}
\label{fig02}
\end{figure}

Our proposed network employs an encoder-decoder architecture backbone (Fig. \ref {fig02}a). The encoder pathway contains five encoder blocks with the first four followed by max pooling. Thus, the downsampling ratio is 16 in total. The decoder pathway is composed of four alternating upsampling and decoder blocks. The upsampling block will be discussed in the next section. The classic UNet employs symmetrical encoder and decoder pathways, where the basic unit (Fig. \ref {fig02}b) is implemented for both the encoder and decoder blocks \cite{Ronneberger2015}. Although this simple design contributes to the efficiency of the network, the effectiveness needs to be explored. Inspired by the recent wide spread usage of ResNet \cite{He2016}, we investigated the feasibility of another two configurations, deep unit (Fig. \ref {fig02}c) and res unit (Fig. \ref {fig02}d).

For the three different units, we have the respective outputs as follows:
\begin{equation}
y_{basic}(x)=\delta(W_{b2}*\delta(W_{b1}*x+b_{b1})+b_{b2})
\label{eq01}
\end{equation}
\begin{equation}
y_{deep}(x)=\delta(W_{d3}*\delta(W_{d2}*\delta(W_{d1}*x+b_{d1})+b_{d2})+b_{d3})
\label{eq02}
\end{equation}
\begin{equation}
y_{res}(x)=\delta((W_{r3}*\delta(W_{r2}*\delta(W_{r1}*x+b_{r1})+b_{r2})+b_{r3}) \\
+(W_{r1}*x+b_{r1}))
\label{eq03}
\end{equation}
where \emph{y} is the respective output of the different units and \emph{x} is the corresponding input. $\delta$ refers to the ReLU function. \emph{W} and \emph{b} refer to the weights and bias of the different convolution layers. * is the convolution operation.

Moreover, we also evaluated the different combinations of applying the three units as the encoder/decoder block. In the results section, we will show that constructing an asymmetrical network backbone by applying the res unit as the encoder block and the basic unit as the decoder block, the network could achieve the best segmentation performance.

\subsection{Attention-guided dense-upsampling block}
\begin{figure}[!t]
\centering
\includegraphics[scale=0.38]{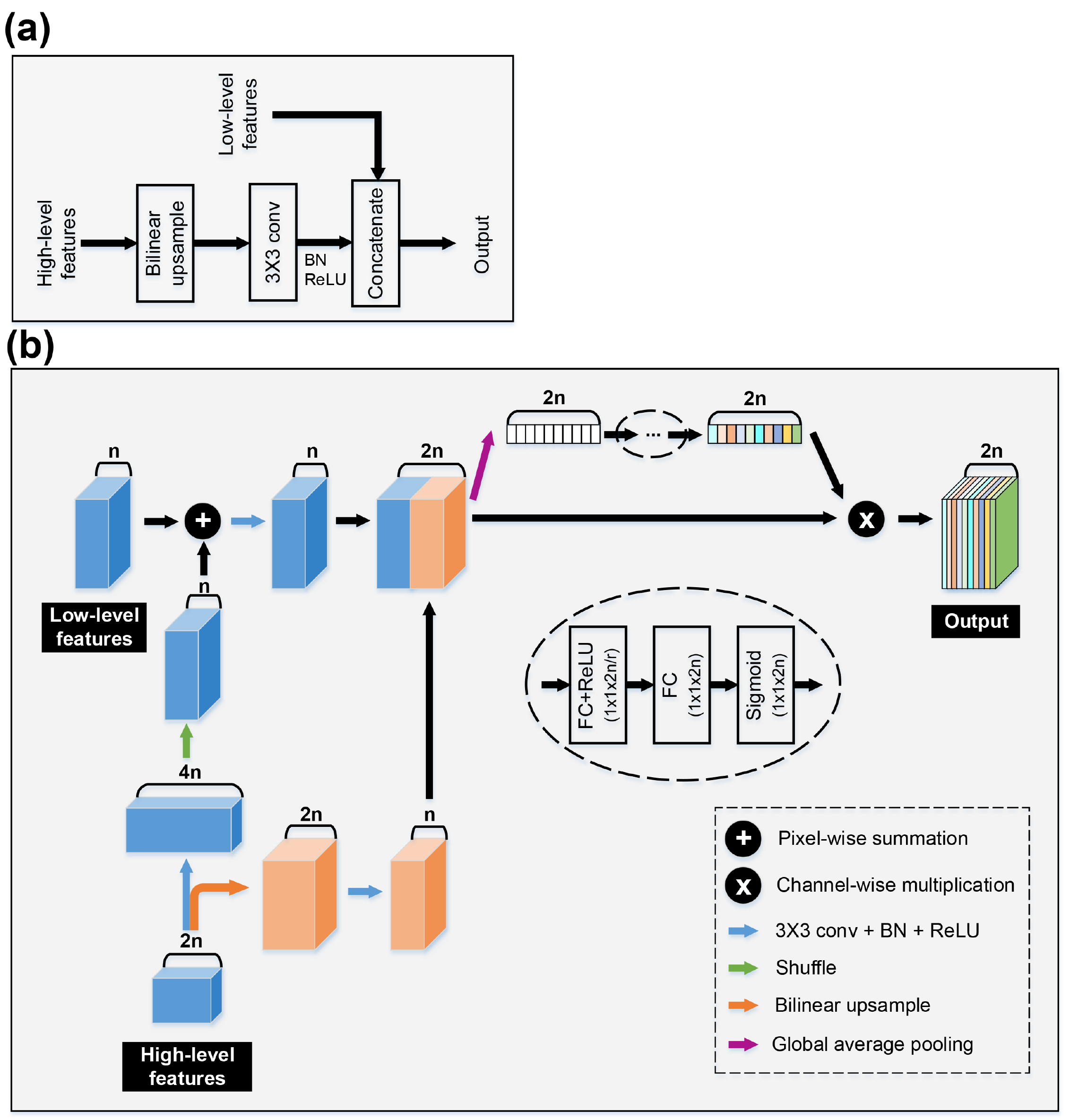}
\caption{Upsampling block. (a) The bilinear upsampling block (BU block). (b) The attention-guided dense-upsampling block (AU block). (For interpretation of the references to color in this figure legend, the reader is referred to the web version of this article.)}
\label{fig03}
\end{figure}

Our major novelty regarding the network design lies in the upsampling block, where we introduce our proposed AU block (Fig. \ref {fig03}b). The original UNet used deconvolution to upsample the feature maps \cite{Ronneberger2015}. However, our preliminary experiments found that deconvolution was not as effective as bilinear upsampling for our application (supplementary file Table S1), and thus, bilinear upsampling was utilized throughout the study.

The bilinear upsampling block (BU block) of UNet is shown in Fig. \ref {fig03}a, where the high-level features are simply upsampled and concatenated with the low-level features after passing a convolution layer. The goal of the proposed AU block (Fig. \ref {fig03}b) is to extract all important information from both high- and low-level features. The high-level low-resolution features ($F_{high}$) are firstly upsampled using two different methods. One is dense upsampling convolution ($F_{duc}$), and the other is bilinear upsampling followed by a convolution layer ($F_{buc}$). The convolution layer is always followed by batch normalization and ReLU activation unless otherwise specified. Then, $F_{duc}$ is combined with the low-level features ($F_{low}$) by summation ($F_{sum}$). A convolution layer is applied before $F_{sum}$ is concatenated with $F_{buc}$ ($F_{concat}$) to smooth the concatenation process. In this way, we expect that $F_{concat}$ contains all the information from both $F_{high}$ and $F_{low}$.

The next step is to select the important information from $F_{concat}$. Motivated by the squeeze-and-excitation networks \cite{Hu2018}, we adopt a channel-wise attention. Firstly, global average pooling is applied to obtain a channel-wise descriptor $Z_{c}$:
\begin{equation}
Z_c=\frac{1}{H \times W}\sum_{i=1}^H\sum_{j=1}^W(F_{concat,c}(i,j))
\label{eq04}
\end{equation}
where \emph{F}$_{concat,c}$ is the \emph($c^{th}$) channel of $F_{concat}$. \emph{H} and \emph{W} refer to the height and width of $F_{concat,c}$. $Z_{c}$ passes through two fully connected layers (FC layers), one with ReLU and one without, and a Sigmoid function to get the channel-wise weights \emph{S}:
\begin{equation}
S=\sigma(W_2*\delta(W_1*Z+b_1)+b_2)
\label{eq05}
\end{equation}
where $\sigma$ refers to the Sigmoid function. $W_1\in R^{2n/r\times2n}$, $W_2\in R^{2n\times2n/r}$, $b_1\in R^{2n/r}$, and $b_2\in R^{2n}$ are the weights and bias of the FC layers, respectively. \emph{r} is a reduction ratio. The output of the AU block is:
\begin{equation}
\tilde{F}_{concat,c}=S_c\cdot Z_c
\label{eq06}
\end{equation}

After that, $\tilde{F}_{concat,c}$ goes through a basic unit (Fig. \ref {fig02}b), which is composed of two convolution layers, and then, is treated as the high-level feature input to the next AU block.

\subsection{Loss function}
The commonly used cross-entropy loss function for two-class segmentation task is defined as:
\begin{equation}
L_{CE}=-\frac{1}{N}(y_i\sum_{i=1}^Np_i+(1-y_i)\sum_{i=1}^N(1-p_i))
\label{eq07}
\end{equation}
For 2D inputs, \emph{N} is the total number of pixels in the image. $y_i\in \{0,1\}$ is the ground truth label of the $i^{th}$ pixel with 0 refers to the background and 1 refers to foreground. $p_i\in [0,1]$ is the corresponding predicted probability of the pixel belonging to the foreground class.

From the definition, positive and negative pixels contribute equally to the cross-entropy loss. However, from Fig. \ref {fig01}, we know a severe class imbalance problem exists for both datasets that masses only occupy small regions of the whole mammograms. Minimization of the cross-entropy loss function may bias the model towards correctly predicting the negative class. To solve this issue, we introduced another loss function, the Dice loss. The Dice loss in our situation is defined as:
\begin{equation}
L_{Dice}=1-\frac{2\sum_{i=1}^Np_iy_i+\varepsilon}{\sum_{i=1}^Np_i+\sum_{i=1}^Ny_i+\varepsilon}
\label{eq08}
\end{equation}
where $\varepsilon$ is a constant to keep numerical stability. It has been reported that applying only the Dice loss makes the optimization process unstable \cite{Zhu2018}. Therefore, we use a combined loss function for our model, which is defined as:
\begin{equation}
L=L_{Dice}+\alpha L_{CE}
\label{eq09}
\end{equation}
where $\alpha$ is a weight constant to control the trade-off between the cross-entropy loss and the dice loss.

\subsection{Evaluation metrics}
To quantitatively evaluate the proposed model, we used dice similarity coefficient (\emph{DSC}), sensitivity (\emph{SEN}), relative area difference ($\Delta A$), and Hausdorff distance (\emph{HAU}) to characterize the performances of the methods on the test datasets. We use the overall average metrics to select the best model during the network architecture optimization. To comprehensively compare our final model to the existing networks, in addition to the overall average metrics, we also evaluate the results with respect to the image properties for the CBIS-DDSM dataset (Fig. \ref{fig01}c-g). \emph{DSC}, \emph{SEN}, $\Delta A$, and \emph{HAU} are defined as:
\begin{equation}
DSC=\frac{2TP}{2TP+FP+FN}
\label{eq10}
\end{equation}
\begin{equation}
SEN=\frac{TP}{TP+FN}
\label{eq11}
\end{equation}
\begin{equation}
\Delta A=\frac{|A_{pred}-A_{GT}|}{A_{GT}}=\frac{|(TP+FP)-(TP+FN)|}{TP+FN}
\label{eq12}
\end{equation}
\begin{equation}
HAU=max(h(pred, GT), h(GT, pred))
\label{eq13}
\end{equation}
where \emph{pred} refers to network predictions and \emph{GT} referes to ground truth segmentations. $A_{pred}$ refers to the predicted mass area and $A_{GT}$ refers to the ground-truth mass area. \emph{TP}, \emph{FP}, and \emph{FN} refer to true positives, false positives, and false negatives. $h(A, B) = max(a\in A)(b\in B)||a-b||$ and $||\cdot||$ refers to the L2 distance between the two points.

Differences between the different models were evaluated by Wilcoxon signed-rank test with a significance threshold of $p < 0.05$.

\subsection{Experimental Set-up}
Our proposed network as well as the comparison models were implemented with PyTorch \cite{Paszke2017}. Network training and testing were run on a NVIDIA GeForce GTX 1080Ti GPU (11G) with batch size of 4. We used ADAM with the AMSGRAD optimization method \cite{Reddi2018}. The learning rate was initially set to $1e^{-4}$, and step decay policy was applied, specifically with [40, 30, 30, 20] epochs at the learning rate of [$1e^{-4}$, $5e^{-5}$, $1e^{-5}$, $1e^{-6}$]. The INbreast dataset contains 107 images, which may limit the proper training of a deep neural network. Therefore, we tried to fine-tune the models pretrained on the CBIS-DDSM dataset. We set the respective hyper-parameters in (\ref {eq08}) and (\ref {eq09}) empirically to $\varepsilon =1.0$ and $\alpha =1.0$. We have tested $\alpha$ with different values (0.5, 1.0 and 2) and found that 1.0 achieved the best segmentation performance (supplementary file Table S2). The determination of the reduction ratio \emph{r} will be discussed in the results section.

To validate the effectiveness of our proposed AUNet, we conducted ablation experiments. Specifically, to select the best network backbone, we have tried to substitute the encoder and decoder blocks in Fig. \ref {fig02}a with the deep unit (Fig. \ref {fig02}c; Deep-UNet) or res unit (Fig. \ref {fig02}d; Res-UNet) but keep the BU block (Fig. \ref {fig03}a) unchanged. In addition, different combinations of the encoder and decoder units have been tested to check the feasibility of symmetric and asymmetric structures. Finally, we compare the segmentation results of the proposed AUNet with three established FCNs, UNet \cite{Ronneberger2015}, FusionNet \cite{Quan2016}, and FCDenseNet \cite{Jegou2017}. The original UNet utilizes deconvolution for upsampling. However, experimental results demonstrated that bilinear upsampling is more effective for our application (supplementary file Table S1). We adopted bilinear upsampling for all the networks. FusionNet introduces residual connections to UNet and increases the network depth by adding more convolution layers in each unit (5 convolutions per unit). FCDenseNet103 extends the recently published architecture DenseNet to fully convolutional networks for image segmentation task. Similarly, all the networks were trained from scratch for the CBIS-DDSM dataset and fine-tuning was investigated on the INbreast dataset. We show that although FusionNet and FCDenseNet103 are much deeper than AUNet, AUNet could still generate better segmentation results, which highlights the effectiveness of the proposed AU block. Three independent experiments were done for each network and the results are presented as ($mean \pm s.d.$).

\section{Experimental results}
In this section, we present the results on the two public datasets, CBIS-DDSM and INbreast, and compare the results of the proposed AUNet to other FCNs.

\subsection{Results on CBIS-DDSM dataset}
In this section, we firstly discuss the choice of the different encoder/decoder blocks. Then the determination of the reduction ratio \emph{r} is demonstrated. Finally, we compare the results of the optimized AUNet to the three FCNs.

\subsubsection{Optimization of the network backbone}
Results of networks employing different encoder and decoder blocks are presented in Table \ref {tab1}. The model names indicate the units applied with the first word referring to the encoder block and the second referring to the decoder block. For example, the model Basic-Deep-UNet means we utilized the basic unit (Fig. \ref {fig02}b) for the encoder pathway and the deep unit (Fig. \ref {fig02}c) for the decoder pathway. From Table \ref {tab1}, two general conclusions could be made: a) Deeper networks generally achieve better performances with higher \emph{DSC}, higher \emph{SEN}, lower $\Delta A$, and lower \emph{HAU} (compare UNet to Deep-Deep-UNet); b) Models with asymmetric structures, especially those employing the basic unit in only one pathway, perform better than models with symmetric structures (compare Res-Basic-UNet to Res-Res-UNet and Res-Deep-UNet).

\begin{table}[!t]
\caption{\label{tab1}Ablation experiments employing different encoder-decoder blocks}
\centering
\small
\begin{tabular}{ccccc}
\toprule[1pt]
Models & \emph{DSC} (\%) & \emph{SEN} (\%) & $\Delta A$ (\%) & \emph{HAU} \\
\midrule[1pt]
UNet (Basic-Basic) & $73.6 \pm 0.2$ & $79.4 \pm 1.3$ & $42.7 \pm 3.1$ & $3.38 \pm 0.04$\\
%\hline
Basic-Deep-UNet & $74.3 \pm 0.1$ & $78.8 \pm 0.4$ & $37.7 \pm 1.2$ & $3.28 \pm 0.05$\\
%\hline
Basic-Res-UNet & $74.6 \pm 0.3$ & $80.0 \pm 0.7$ & $42.0 \pm 1.6$ & $3.31 \pm 0.07$\\
\midrule[0.5pt]
Deep-Basic-UNet & $77.3 \pm 0.5$ & $\color{red}82.5 \pm 0.8$ & $36.6 \pm 0.2$ & $\color{red}3.16 \pm 0.03$\\
%\hline
Deep-Deep-UNet & $77.3 \pm 0.4$ & $81.3 \pm 1.2$ & $35.7 \pm 1.8$ & $3.19 \pm 0.04$\\
%\hline
Deep-Res-UNet & $76.9 \pm 0.7$ & $81.6 \pm 1.1$ & $36.4 \pm 1.9$ & $\color{red}3.16 \pm 0.05$\\
\midrule[0.5pt]
\textbf{Res-Basic-UNet} & $\color{red}77.5 \pm 0.2$ & $\color{blue}82.3 \pm 1.5$ & $\color{blue}35.2 \pm 3.6$ & $\color{blue}3.18 \pm 0.04$\\
%\hline
Res-Deep-UNet & $76.3 \pm 0.2$ & $81.9 \pm 1.5$ & $38.4 \pm 3.8$ & $3.23 \pm 0.04$\\
%\hline
Res-Res-UNet & $76.2 \pm 0.1$ & $80.4 \pm 0.2$ & $\color{red}34.7 \pm 0.9$ & $3.19 \pm 0.02$\\
\bottomrule[1pt]
\end{tabular}
\end{table}

By taking all the four evaluation parameters into consideration, we selected the model ‘Res-Basic-UNet’ as our network backbone since it achieves the highest average \emph{DSC} ($0.775 \pm 0.002$) among all the models and, in the meantime, comparable \emph{SEN} ($0.823 \pm 0.015$ vs. $0.825 \pm 0.008$), $\Delta A$ ($0.352 \pm 0.036$ vs. $0.347 \pm 0.009$), and \emph{HAU} ($3.18 \pm 0.04$ vs. $3.16 \pm 0.03$) to the respective best results.

\subsubsection{Performance enhancement by the AU block}
The introduction of the AU block (Fig. \ref {fig03}b) to our network backbone brings an obvious performance increment shown by all the four evaluation characteristics (Table \ref {tab2}). The reduction ratio \emph{r} is very important for the capacity and computational cost of the proposed AUNet. Therefore, we have conducted experiments to finalize the selection. A wide range of \emph{r} has been tested from 2 to 32. Results indicate that with $r=16$, the best model performance could be achieved (Table \ref {tab2}). Besides, it could also be observed that regardless of the choice of \emph{r}, the proposed AU Block could always enhance the segmentation performance compared to the selected network backbone (Res-Basic-UNet), which demonstrates the general effectiveness of the proposed block. For all the following experiments, $r=16$ is applied unless otherwise specified.

\begin{table}[!t]
\caption{\label{tab2}Investigating the influence of the reduction ratio \emph{r}}
\centering
\small
\begin{tabular}{ccccc}
\toprule[1pt]
Reduction ratio & \emph{DSC} (\%) & \emph{SEN} (\%) & $\Delta A$ (\%) & \emph{HAU}\\
\midrule[1pt]
2 & $80.6 \pm 0.2$ & $83.7 \pm 0.9$ & $29.1 \pm 1.2$ & $3.02 \pm 0.03$\\
4 & $80.8 \pm 0.2$ & $84.6 \pm 0.4$ & $28.5 \pm 0.8$ & $2.97 \pm 0.02$\\
8 & $81.0 \pm 0.3$ & $84.4 \pm 0.3$ & $29.1 \pm 2.4$ & $2.97 \pm 0.06$\\
\textbf{16} & $\color{red}{81.8 \pm 0.0}$ & $\color{red}{84.9 \pm 0.3}$ & $\color{red}{26.9 \pm 0.3}$ & $\color{red}{2.96 \pm 0.03}$\\
32 & $80.8 \pm 0.0$ & $84.1 \pm 0.5$ & $28.5 \pm 1.4$ & $2.98 \pm 0.01$\\
\bottomrule[1pt]
\end{tabular}
\end{table}

\subsubsection{Comparison to established FCNs}
Our proposed AUNet achieves the best segmentation results when compared to established FCNs (Table \ref {tab3}). Comparing among the three established models, FusionNet gives the highest \emph{DSC}, the lowest $\Delta A$, and the lowest \emph{HAU} whereas FCDenseNet103 presents the highest \emph{SEN}. This indicates that FCDenseNet103 increases its capability of finding the mass locations by generating more false positives. Since FCDenseNet103 is much deeper than the other networks, it suggests that very deep networks perform worse on the mammographic datasets probably caused by overfitting. On the other hand, our proposed AUNet achieves the best results with the highest \emph{DSC}, the highest \emph{SEN}, the lowest $\Delta A$, and the lowest \emph{HAU}, which demonstrates the superiority and robustness of our proposed network. Our model shows an average \emph{DSC} increase of at least 2\% (statistically significant with $p < 0.05$ by Wilcoxon signed-rank test), \emph{SEN} increase of 0.7\%, $\Delta A$ decrease of 4.4\%, and \emph{HAU} decrease of 0.05 compared to the respective best performed FCNs.

\begin{table}[!t]
\caption{\label{tab3}Segmentation performance of different FCNs on the CBIS-DDSM dataset}
\centering
\small
\begin{tabular}{ccccc}
\toprule[1pt]
Models & \emph{DSC} (\%) & \emph{SEN} (\%) & $\Delta A$ (\%) & \emph{HAU}\\
\midrule[1pt]
UNet & $73.6 \pm 0.2$ & $79.4 \pm 1.3$ & $42.7 \pm 3.1$ & $3.38 \pm 0.04$\\
FusionNet & $79.8 \pm 0.5$ & $83.9 \pm 0.8$ & $31.3 \pm 0.5$ & $3.01 \pm 0.03$\\
FCDenseNet103 & $78.2 \pm 0.1$ & $84.2 \pm 0.6$ & $40.2 \pm 0.3$ & $3.13 \pm 0.04$\\
\midrule[0.5pt]
\textbf{AUNet} & $\color{red}{81.8 \pm 0.0}$ & $\color{red}{84.9 \pm 0.3}$ & $\color{red}{26.9 \pm 0.3}$ & $\color{red}{2.96 \pm 0.03}$\\
\bottomrule[1pt]
\end{tabular}
\end{table}

Considering the inherent differences among the images having different categories (subtlety, BIRADS, mass shape, mass margin, and pathology), the segmentation performances of the different networks are also presented with regards to these properties. Combining the different categories (21 in total: 5 subtlety groups, 4 BIRADS categories, 5 shape groups, 5 margin categories, and 2 pathology groups) with the different evaluation metrics (\emph{DSC}, \emph{SEN}, $\Delta A$, and \emph{HAU}), there are 84 cases (detalied results in supplementary file Table S3-S6). Overall, our AUNet still achieves the best results, ranking the $1^{st}$ in 56 cases (16 for \emph{DSC}, 11 for \emph{SEN}, 14 for $\Delta A$, and 15 for \emph{HAU}). FusionNet and FCDenseNet103 obtain the best results in 15 and 10 cases, respectively. UNet performs the worst in this aspect with only 3 $1^{st}$ cases.

\begin{figure}[!t]
\centering
\includegraphics[scale=0.7]{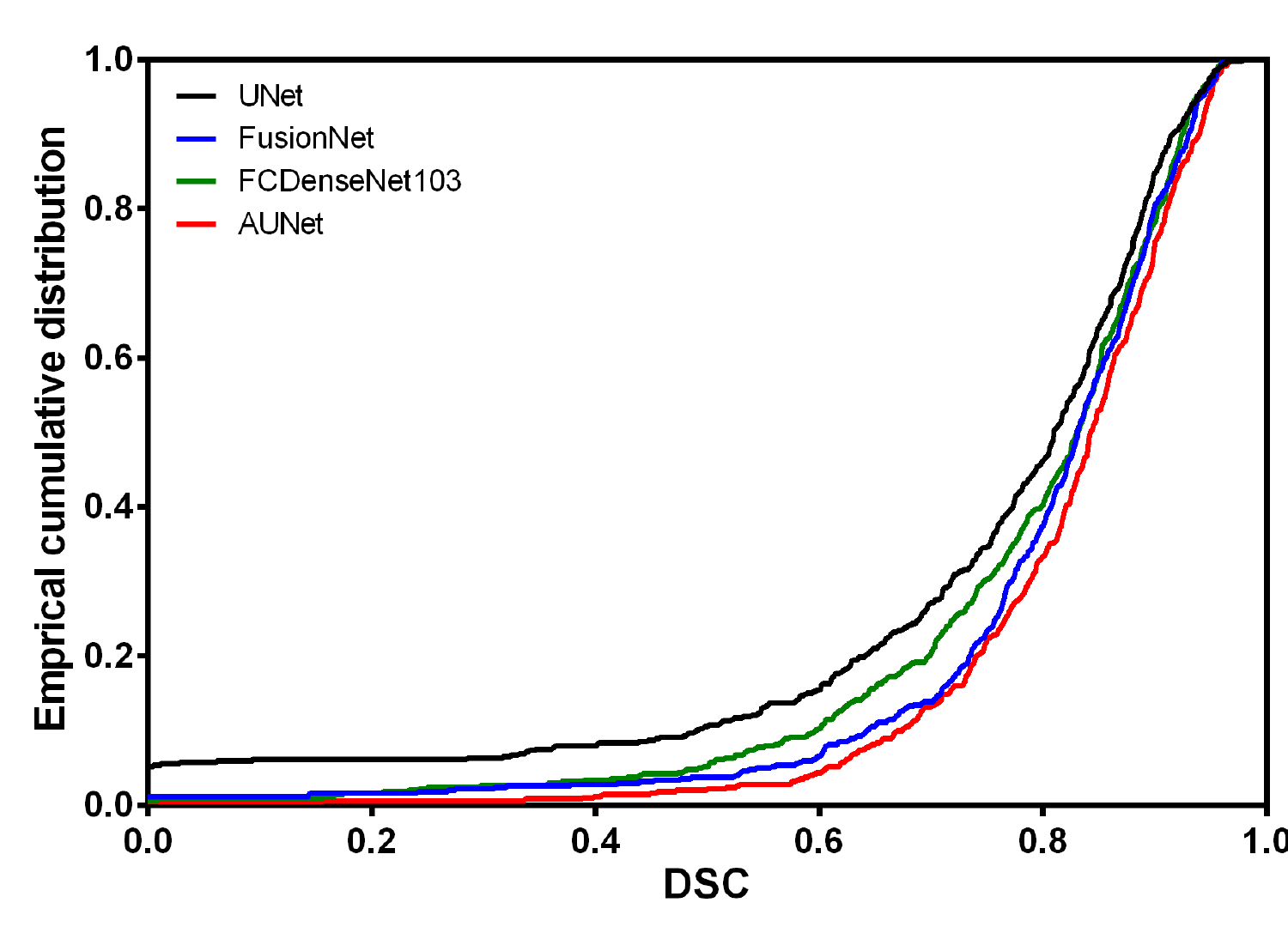}
\caption{Empirical cumulative distribution of \emph{DSC} on CBIS-DDSM. (For interpretation of the references to color in this figure legend, the reader is referred to the web version of this article.)}
\label{fig04}
\end{figure}

To directly compare the performances of the different networks, the empirical cumulative distributions of \emph{DSC} were plotted (Fig. \ref {fig04}). The closer the distribution line to the lower right position in the figure, the more images are segmented with high \emph{DSC} values by the corresponding network. Thus, we could conclude that for the CBIS-DDSM dataset, AUNet achieves the best segmentation performance, followed by FusionNet, FCDenseNet, and UNet.

\subsection{Results on INbreast dataset}
The INbreast dataset is smaller than the CBIS-DDSM dataset. As such, we tried to re-use the CBIS-DDSM trained models and fine-tuned those models using the INbreast dataset. Moreover, 5-fold cross-validation experiments were conducted to generate meaningful and convincing results.

\begin{figure}[!t]
\centering
\includegraphics[scale=0.7]{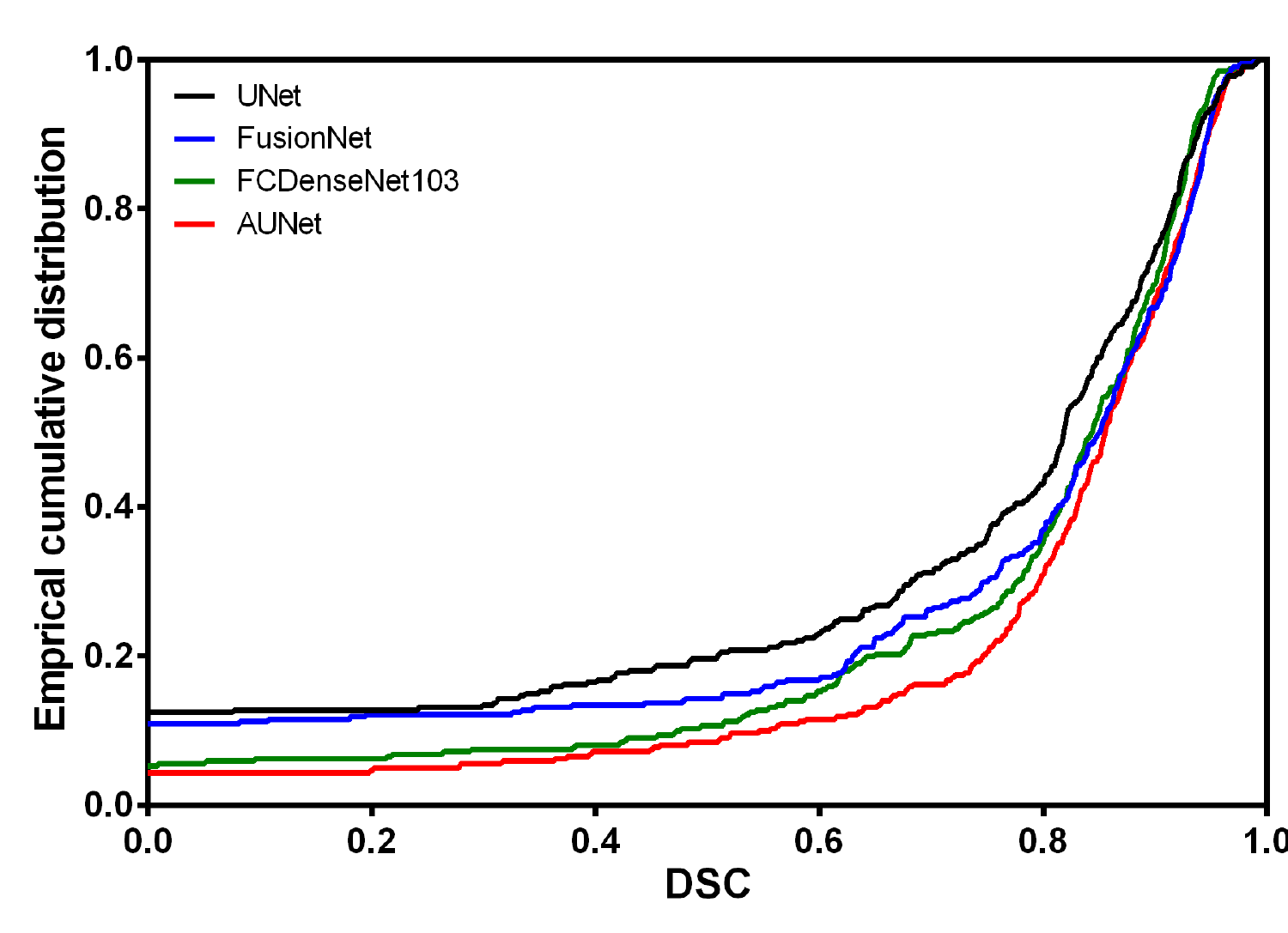}
\caption{Empirical cumulative distribution of \emph{DSC} on INbreast with pretraining. (For interpretation of the references to color in this figure legend, the reader is referred to the web version of this article.)}
\label{fig05}
\end{figure}

The segmentation results of the proposed AUNet and the three established models with/without pretraining on CBIS-DDSM are listed in Table \ref {tab4}. It could be observed that with or without the pretraining step, AUNet always generates the best segmentation results and pretraining improves the segmentation performance of all the methods significantly. With pretraining on CBIS-DDSM, the results of the three established models present a different pattern from the CBIS-DDSM dataset. Among the three established models, FCDenseNet103 generates the highest \emph{DSC} and \emph{SEN} value, UNet shows the lowest $\Delta A$, and FusionNet gives the lowest \emph{HAU}. It is interesting that FusionNet shows much worse performance on INbreast than that on CBIS-DDSM. On the other hand, compared to the three models, our proposed AUNet still gives the best segmentation results with the highest \emph{DSC}, the highest \emph{SEN}, the lowest $\Delta A$, and the lowest \emph{HAU}. AUNet shows an average \emph{DSC} increase of at least 3\% (statistically significant with $p < 0.05$ by Wilcoxon signed-rank test), \emph{SEN} increase of 2.9\%, $\Delta A$ decrease of 6.5\%, and \emph{HAU} decrease of 0.29 (statistically significant with $p < 0.05$ by Wilcoxon signed-rank test). Similarly, the empirical cumulative distribution plot indicates that for INbreast, AUNet still achieves the best segmentation performance, followed by FCDenseNet, FusionNet, and UNet (Fig. \ref {fig05}).

\begin{table}[!t]
\caption{\label{tab4}Segmentation performance of different FCNs on the INbreast dataset}
\centering
\small
\begin{tabular}{ccccc}
\toprule[1pt]
Models & \emph{DSC} (\%) & \emph{SEN} (\%)& $\Delta A$ (\%) & \emph{HAU} \\
\midrule[1pt]
UNet (w/o $^*$) & $62.3 \pm 3.7$ & $62.7 \pm 4.0$ & $54.3 \pm 19.7$ & $4.73 \pm 0.26$\\
UNet (w/ $^{\#}$) & $69.3 \pm 6.8$ & $70.4 \pm 8.8$ & $44.0 \pm 13.3$ & $4.54 \pm 0.42$\\
FusionNet (w/o) & $62.1 \pm 5.8$ & $65.1 \pm 5.4$ & $62.7 \pm 30.9$ & $4.80 \pm 0.33$\\
FusionNet (w/) & $73.2 \pm 5.8$ & $74.6 \pm 5.4$ & $69.8 \pm 33.8$ & $4.33 \pm 0.34$\\
FCDenseNet103 (w/o) & $42.9 \pm 8.5$ & $52.8 \pm 11.9$ & $149.5 \pm 71.8$ & $6.20 \pm 0.52$\\
FCDenseNet103 (w/) & $76.1 \pm 4.6$ & $77.9 \pm 4.7$ & $47.1 \pm 17.3$ & $4.35 \pm 0.35$\\
\midrule[0.5pt]
AUNet (w/o) & $64.0 \pm 7.6$ & $66.0 \pm 7.4$ & $51.6 \pm 21.0$ & $4.66 \pm 0.43$\\
\textbf{AUNet (w/)} & $\color{red}{79.1 \pm 6.0}$ & $\color{red}{80.8 \pm 7.1}$ & $\color{red}{37.6 \pm 15.4}$ & $\color{red}{4.04 \pm 0.33}$\\
\bottomrule[1pt]
\multicolumn{5}{@{}l}{$^*$w/o--Without pretraining on CBIS-DDSM} \\
\multicolumn{5}{@{}l}{$^{\#}$w/--With pretraining on CBIS-DDSM}
\end{tabular}
\end{table}

\subsection{Qualitative results}
\begin{figure*}[!t]
\centering
\includegraphics[scale=0.98]{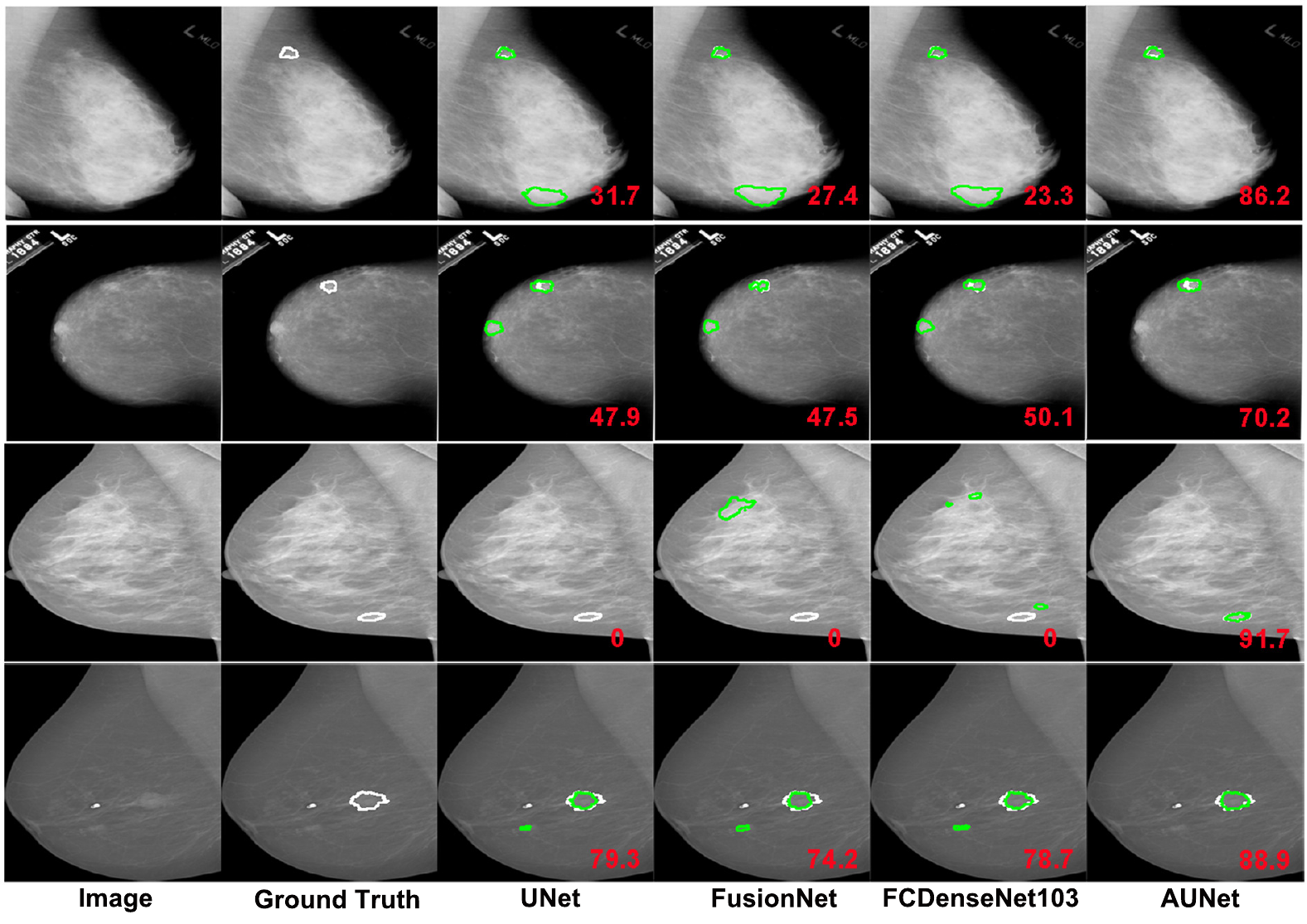}
\caption{Segmentation results of different networks. From left to right, the columns correspond to the input images, the ground truth labels, the segmentation results of UNet, FusionNet, FCDenseNet103, and our proposed AUNet, respectively. The white circles indicate the boundaries of the labels and the green circles indicate the boundaries of the segmentation results. The red number on the right bottom of each image is the \emph{DSC} value of the segmentation result. The first two rows are from the CBIS-DDSM dataset and the last two rows are from the INbreast dataset. (For interpretation of the references to color in this figure legend, the reader is referred to the web version of this article.)}
\label{fig06}
\end{figure*}

Fig. \ref {fig06} presents several segmentation results generated by the different networks for qualitative comparisons. We can see, overall, our proposed AUNet performs better than the other three FCNs for our whole mammographic mass segmentation task. In addition, it could be observed that AUNet displays an impressive ability to suppress the false positive results of UNet without increasing the number of false negatives, whereas both FusionNet and FCDenseNet103 are not effective in this aspect or even make the situation worse (Fig. \ref {fig06}; the first, second, and last rows). This observation is consistent with the quantitative results discussed before. Lastly, our AUNet could give accurate segmented masses for difficult samples when the other three networks could barely find the targeted regions at all, such as the third example in Fig. \ref {fig06}.

\subsection{Results on extracted image patches}
In order to compare the performance of proposed network directly to the literature on breast mass segmentation, we also conducted experiments on extracted mass-centered image patches for the INbreast dataset. For each mammogram, we first found the smallest rectangular that could accommodate the mass. Then, the mass-centered image patch was extracted through enlarging the rectangular by 20\% in area with an equal elongation ratio in width and height of $\sqrt{1.2}$. Similar to the whole mammogram situation, 5-fold cross-validation experiments with three replicates were done. Results in Table \ref {tab5} confirms that our proposed AUNet could also achieve the best segmentation results on mass-centered image patches compared to both the three FCNs and the literature reported results.

\begin{table}[!t]
\caption{\label{tab5}Segmentation performance on mass-centered image patches for INbreast dataset}
\centering
\small
\begin{tabular}{ccccc}
\specialrule{1pt}{2pt}{2pt}
Models & \emph{DSC} (\%) & \emph{SEN} (\%) & $\Delta A$ (\%) & \emph{HAU}\\
\specialrule{1pt}{2pt}{10pt}
Cardoso et al., 2015 (\cite{Cardoso2015}) & $0.88\times100\%$ & $-$ & $-$ & $-$\\
\specialrule{0pt}{10pt}{2pt}
Dhungel et al., 2015b (\cite{Dhungel2015b}) & $(0.90 \pm 0.06)\times100\%$ & $-$ & $-$ & $-$ \\
\specialrule{0pt}{2pt}{2pt}
Dhungel et al., 2017 (\cite{Dhungel2017})$^*$ & $(0.85 \pm 0.02)\times100\%$ & $-$ & $-$ & $-$ \\
\specialrule{0.5pt}{2pt}{2pt}
UNet & $92.0 \pm 0.8$ & $93.1 \pm 1.2$ & $8.3 \pm 2.5$ & $6.88 \pm 0.18$\\
FusionNet & $92.0 \pm 0.8$ & $92.7 \pm 1.0$ & $8.1 \pm 2.9$ & $6.94 \pm 0.19$\\
FCDenseNet103 & $89.5 \pm 0.8$ & $89.6 \pm 2.0$ & $11.7 \pm 2.2$ & $7.23 \pm 0.14$\\
\midrule[0.5pt]
\textbf{AUNet} & $\color{red}{92.4 \pm 0.9}$ & $\color{red}{93.7 \pm 0.9}$ & $\color{red}{7.5 \pm 2.6}$ & $\color{red}{6.85 \pm 0.28}$\\
\bottomrule[1pt]
\multicolumn{4}{@{}l}{$^*$Patches were extracted based on detection results.} \\
\end{tabular}
\end{table}

\begin{table}[!t]
\caption{\label{tab6}Computational complexities of different networks}
\centering
\small
\begin{tabular}{ccccc}
\specialrule{1pt}{2pt}{2pt}
Models & UNet & FusionNet & FCDenseNet103 & AUNet (R=16) \\
\specialrule{1pt}{2pt}{10pt}
Convolutional and FC layers & 23 & 50 & 103 & 44 \\
\specialrule{0pt}{2pt}{2pt}
Parameters (million) & 34.5 & 78.5 & 13.9 & 75.5 \\
\specialrule{0pt}{2pt}{2pt}
FPS (with $256 \times 256$ inputs) & 59 & 36 & 27 & 32 \\
\bottomrule[1pt]
\end{tabular}
\end{table}

\subsection{Model complexity}
Table \ref {tab6} lists the total number of convolutional and FC layers, the optimizable parameters, and the inference time in terms of frames per second (FPS) with input resized to $256 \times 256$. Obviously, UNet is the simplest and fastest model, and the other three models (FusionNet, FCDenseNet103, and AUNet) have similar inference speeds with AUNet achieves the best segmentation performance.

\section{Discussion}
\begin{figure}[!t]
\centering
\includegraphics[scale=1.3]{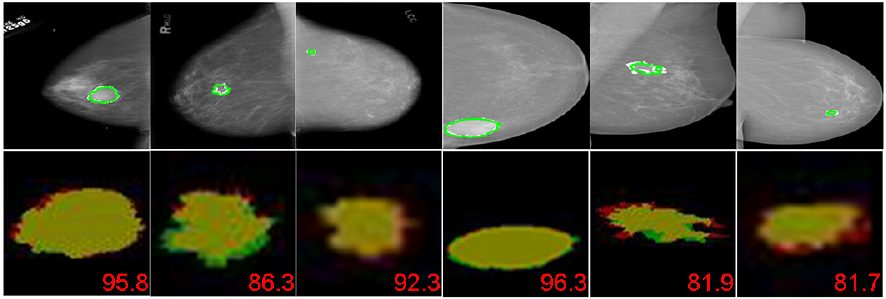}
\caption{Segmentation results of AUNet when input images contain masses of different shapes and sizes. The left three columns are images from the CBIS-DDSM dataset and the right three columns are from the INbreast dataset. The white circles indicate the boundaries of the labels and the green circles indicate the boundaries of the segmentation results. Red color regions indicate the ground-truth masses and green indicate the segmentation results. Yellow color regions indicate the overlap between the segmentation and the ground-truth. The red number on the right bottom of each image is the \emph{DSC} value of the segmentation result. (For interpretation of the references to color in this figure legend, the reader is referred to the web version of this article.)}
\label{fig07}
\end{figure}

Segmentation of mammographic masses is a challenging task as mammograms have low signal-to-noise ratio and breast masses may vary in shapes and sizes. An easy alternative is to segment masses from extracted ROIs. However, manual extraction of ROIs is a tedious task. Automatic detection algorithms still subject to high false positives and specially designed post processing methods are required to achieve expected performance \cite{Dhungel2015a}. Therefore, automatic breast mass segmentation in whole mammograms is of great clinical value. There are several reports targeting at developing deep learning models for whole mammographic mass segmentation, such as the ASPP-FC-DenseNet and the Attention Dense-U-Net \cite{Jinjin2019,Li2019b}. ASPP-FC-DenseNet achieved a Dice similarity coefficient of 76.97\% on the private dataset and Attention Dense-U-Net achieved a sensitivity of 77.89\% on the selected DDSM dataset. Both are much smaller than the results achieved in this study, which confirms the superiority of our proposed AUNet. Fig. \ref {fig07} presents a few segmentation results of AUNet. We admit that compared to inputs with irregular or small masses, AUNet performs slightly better for inputs with large and regular masses. However, Fig. \ref {fig06} indicates that AUNet still performs better than the three FCNs for inputs with small and irregular masses. Overall, Fig. \ref {fig06} and \ref {fig07} conclude that for inputs with different mass shapes and sizes, our AUNet could always give very accurate segmentation results.

Mammograms are taken with high resolutions. Images from CBIS-DDSM dataset have a width ranging from 1786 to 5431 pixels and a height ranging from 3920 to 6931 pixels. Images form INbreast have either $3328 \times 4084$ or $2560 \times 3328$ pixels. To facilitate the training and testing of deep neural networks, necessary image preprocessing steps are required, such as image patch extraction or resizing. Although patch extraction method can preserve all the original image information and researchers have developed elegant approaches to extract informative image patches \cite{Qin2018}, we adopted resizing in this study. On one hand, it has been suggested in the computer vision field that global contextual information is important for accurate image segmentation \cite{Wang2018b}. Patch extraction restricts the field of view of the network, which may influence the segmentation performance. Therefore, the correlations between the patches need to be carefully considered, which we will investigate in the following work. On the other hand, after resizing, most masses still occupy hundreds to thousands of pixels. We believe these downsampled masses are large enough to preserve the overall mass information. Moreover, different input settings have been tested with gray or RGB inputs, with different resolutions ($256 \times 256$ inputs or $512 \times 512$ inputs), and with fixed aspect ratios by zero padding the images before resizing (Table \ref {tab7}). Although different inputs show influence on the final segmentation results, our proposed AUNet always achieves the best performance (more results in supplementary file Table S7-S9). Thus, it can be anticipated that our method should also be able to achieve the best segmentation performance if the full resolution inputs are utilized. With detailed inspection, the results show that RGB inputs could improve the segmentation performance. Even though it was not investigated in the current study, RGB inputs can also facilitate the direct transfer learning of networks trained on natural images. Resizing to the higher resolution ($512 \times 512$ pixels) showed negative effects on the segmentation performance, which was also observed for the three established FCNs (supplementary file Table S8). This weakened performance might be caused by two reasons. One is due to the GPU memory limitation, batch size of 2 was applied for inputs with $512 \times 512$ pixels instead of 4 for inputs with $256 \times 256$ pixels. The other is it is difficult to accurately define the mass boundaries in mammograms. At higher resolutions, the images are more sensitive to manual label errors. Zero padding brings large regions of background to the inputs and hinders the segmentation process. In this study, our experiments were done with RGB inputs resized to $256 \times 256$ pixels to maximize the segmentation performance. 

\begin{table}[!t]
\caption{\label{tab7}Segmentation performance of AUNet with different input settings on CBIS-DDSM dataset}
\centering
\small
\begin{tabular}{ccccc}
\toprule[1pt]
Inputs & \emph{DSC} (\%) & \emph{SEN} (\%) & $\Delta A$ (\%) & \emph{HAU}\\
\midrule[1pt]
\multirow{2}{4cm}{\centering Gray input \\ (resize $256 \times 256$)} & \multirow{2}{2cm}{\centering $80.9 \pm 0.4$ } & \multirow{2}{2cm}{\centering $84.5 \pm 0.5$} & \multirow{2}{2cm}{\centering $29.9 \pm 0.2$}  & \multirow{2}{2cm}{\centering $2.97 \pm 0.01$} \\ \\ 
\multirow{2}{4cm}{\centering \textbf{RGB input} \\ \textbf{(resize $\bm {256 \times 256}$)}} & \multirow{2}{2cm}{\centering \bm{$81.8 \pm 0.0$} } & \multirow{2}{2cm}{\centering \bm{$84.9 \pm 0.3$}} & \multirow{2}{2cm}{\centering \bm{$26.9 \pm 0.3$}} & \multirow{2}{2cm}{\centering $2.96 \pm 0.03$}\\ \\ 
\multirow{2}{4cm}{\centering RGB input \\ (resize $512 \times 512$)} & \multirow{2}{2cm}{\centering $78.7 \pm 0.5$ } & \multirow{2}{2cm}{\centering $81.1 \pm 0.1$} & \multirow{2}{2cm}{\centering $28.6 \pm 1.0$} & \multirow{2}{2cm}{\centering $4.30 \pm 0.02$} \\ \\ 
\multirow{3}{4cm}{\centering RGB input \\ (pad \& resize \\ $256 \times 256$)} & \multirow{3}{2cm}{\centering $78.9 \pm 0.2$ } & \multirow{3}{2cm}{\centering $83.3 \pm 0.2$} & \multirow{3}{2cm}{\centering $30.4 \pm 1.1$} & \multirow{3}{2cm}{\centering \bm{$2.20 \pm 0.01$}}\\ \\ \\ \bottomrule[1pt]
\end{tabular}
\end{table}

\begin{figure}[!t]
\centering
\includegraphics[scale=0.98]{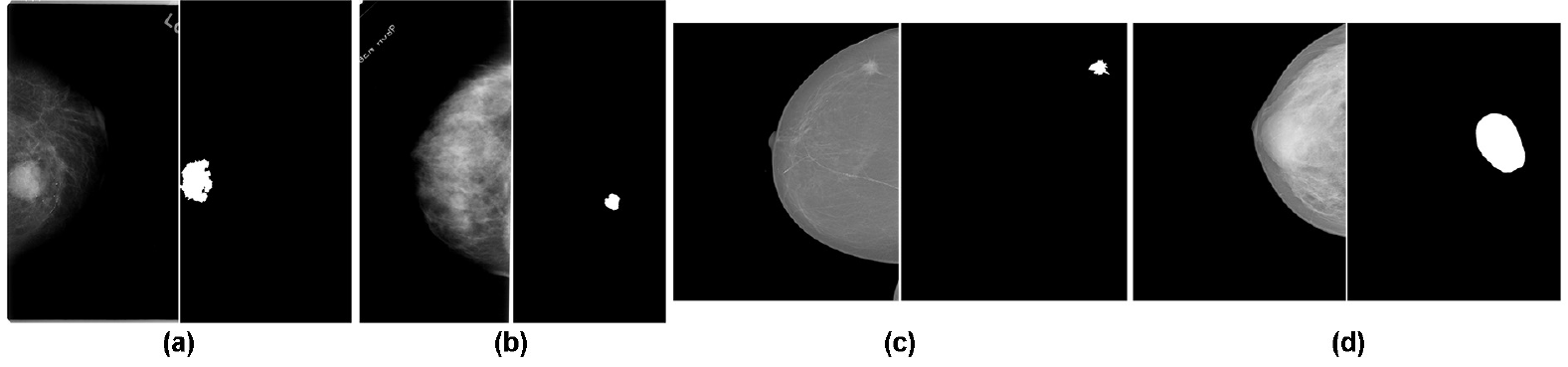}
\caption{Example images from the two datasets. Image and label from the CBIS-DDSM categorized as BIRADS 1 (a), from the CBIS-DDSM categorized as BIRADS 4 (b), from the INbreast categorized as BIRADS 1 (c), and from the INbreast categorized as BIRADS 4 (d).}
\label{fig08}
\end{figure}

Our AUNet, as well as the three comparison networks, showed severely worse performance on the INbreast dataset compared to that on the CBIS-DDSM dataset when trained from scratch (Table \ref{tab3} and \ref{tab4}). A major cause could be the large difference in the sample size. Much better results were obtained when the networks were pretrained on the CBIS-DDSM dataset. But still, the performance is not as good as that on the CBIS-DDSM dataset. Except the sample size, another observable difference between the two datasets is the different image contrasts (Fig. \ref{fig08}). CBIS-DDSM images have a higher contrast in the breast regions than the INbreast images. Although intensity normalization was conducted before the images were inputted into the networks, the differences in the original image contrast might also affect the results. Besides, as shown in Fig. \ref{fig01}, the image distributions are also different between the two datasets, which might influence the results a little bit.

UNet is a very powerful network for biomedical image segmentation \cite{Ronneberger2015} and is the template for many following-up studies \cite{Balagopal2018,Li2019}. Our proposed AUNet adopts a similar encoder-decoder architecture. To enhance the performance, we first investigated the network backbone design. Compared to the basic unit (Fig. \ref {fig02}b) used in both the encoder and decoder pathways of naive UNet, we found that our asymmetrical network backbone Res-Basic-UNet was more suitable for our application. This is reasonable as the res unit (Fig. \ref {fig02}d) in the encoder pathway promotes the information and gradient propagation while the basic unit (Fig. \ref {fig02}b) in the decoder pathway better preserves important semantic information of the high-level features. Our results show that Res-Basic-UNet improves the \emph{DSC} by 3.9\% over UNet.

Then, we believe that the simple bilinear upsampling method and the feature fusion through concatenation adopted by UNet are not effective enough. Significant information loss might happen, which could greatly worsen the segmentation results. Therefore, we proposed a new upsampling block, AU block, to solve these problems. AU block utilizes the high-level features in two means. In one way, the high-level features are densely upsampled and fused with the low-level features by summation. In the other, the high-level features are bilinear upsampled and concatenated with the convolution smoothed summation (Fig. \ref {fig03}b). Moreover, in order to select the rich-informative channels, a channel-wise attention component is used after the concatenation. With AU block, our AUNet increases the \emph{DSC} by another 4.3\% over Res-Basic-UNet. Besides, AUNet outperforms the three widely used FCN segmentation networks and recently by a large margin for both CBIS-DDSM and INbreast datasets.

False positive and false negative are important issues that need to be considered for CAD systems. False positive is commonly found to be the problem that hinders the application of automatic detection algorithms to medical imaging \cite{Dhungel2015a,Samala2016}. It can bring huge psychological stress and depression to patients and result in unnecessary biopsies. False negative, on the other hand, is detrimental for clinical applications which can miss early diagnosis. It is important to reduce both false positive and false negative results. The low signal-to-noise ratio of a mammogram makes it difficult to clearly differentiate the masse from the normal breast tissues (Fig. \ref {fig01}a and Fig. \ref {fig06}). All the three FCNs show serious false positive segmentation results, which greatly affected the evaluation metrics (Fig. \ref {fig06}). On the contrary, AUNet is able to effectively reduce the false positive incidences without increasing the false negative results through the information selection by channel-wise attention. Moreover, thanks to the full utilization of the feature map information, AUNet also performs better at decreasing the false negative results (Fig. \ref {fig06}; the third example).

Breast masses are significant contributors to breast cancers \cite{Giger2013}. Mass segmentation is an important step for the following disease diagnosis and treatment planning. After the mass segmentation, image features can be extracted from and surrounding the specific regions and different analyses can be conducted. These image features could be used to differentiate breast cancer subtypes \cite{Wu2017a}. They were found to be associated with tumor-infiltrating lymphocytes in breast cancer, which is a promising predictive biomarker for the effectiveness of immunotherapy treatment \cite{Wu2018a}. Some of them were identified as valuable prognostic markers for adjuvant and neoadjuvant chemotherapies \cite{Wu2017b,Wu2018b}. As a necessary next step for our current work, we will study the corresponding image feature extraction methods as well as imaging-based disease diagnosis and treatment plan selection in the future.

\section{Conclusion}
In this work, we propose a new network, AUNet, for the mass segmentation in whole mammograms. Specifically, we utilized an asymmetrical encoder-decoder architecture and introduced a new upsampling block, AU block, to boost the segmentation performance. Comprehensive experiments have been conducted. AUNet presented improved segmentation behaviors on both CBIS-DDSM and INbreast datasets compared to existing FCN models, which proves its effectiveness and robustness. In addition, AUNet could greatly reduce both false negative and false positive results. We will make our code available, by which we hope our work can attract and inspire more following-up studies in the field.

\section*{Acknowledgments}
This work was supported by funding from the National Natural Science Foundation of China (61601450, 61871371, and 81830056), Science and Technology Planning Project of Guangdong Province (2017B020227012), the Department of Science and Technology of Shandong Province (2015ZDXX0801A01), and the Fundamental Research Funds of Shandong University (2015QY001 and 2017CXGC1502).

\section*{References}

\bibliographystyle{unsrt}
\bibliography{refs_resub}

\end{document}